\documentclass[conference]{IEEEtran}
\IEEEoverridecommandlockouts

\usepackage{cite}
\usepackage{tabularx}
\usepackage{amsmath,amssymb,amsfonts}
\usepackage{algorithmic}
\usepackage{graphicx}
\usepackage{textcomp}
\usepackage{xcolor}
\def\BibTeX{{\rm B\kern-.05em{\sc i\kern-.025em b}\kern-.08em
    T\kern-.1667em\lower.7ex\hbox{E}\kern-.125emX}}
    
\usepackage{hyperref,tablefootnote,footnotehyper}
\newcommand{\overbar}[1]{\mkern 3mu\overline{\mkern-3mu#1\mkern-3mu}\mkern 3mu}
\newcommand{\thickhat}[1]{\mathbf{\hat{\text{$#1$}}}}
\newcommand{\thickbar}[1]{\mathbf{\bar{\text{$#1$}}}}

\newcommand{\bbC}{{\mathbb{C}}}
\newcolumntype{a}{>{\hsize=1.15\hsize}X}
\newcolumntype{s}{>{\hsize=0.85\hsize}X}
\newcolumntype{b}{>{\hsize=1.3\hsize}X}
\newcolumntype{t}{>{\hsize=0.7\hsize}X}
\newcolumntype{P}[1]{>{\centering\arraybackslash}p{#1}}
\newcolumntype{L}[1]{>{\arraybackslash}p{#1}}
\DeclareMathOperator{\Tr}{Tr}
    
\renewcommand{\thesection}{\Roman{section}}
\renewcommand{\thesubsection}{\Alph{subsection}}
\renewcommand{\thesubsubsection}{\arabic{subsubsection}}
\makeatletter
\renewcommand{\p@subsection}{\thesection.}
\renewcommand{\p@subsubsection}{\thesection.\thesubsection.}
\makeatother

\begin{document}
\bstctlcite{IEEEexample:BSTcontrol}

\title{Subspace Perturbation Analysis for \\ Data-Driven Radar Target Localization}

\author{\IEEEauthorblockN{Shyam Venkatasubramanian}
\IEEEauthorblockA{\textit{Duke University}\\
Durham, NC, USA \\
sv222@duke.edu} \\
\IEEEauthorblockN{Ali Pezeshki}
\IEEEauthorblockA{\textit{Colorado State University}\\
Fort Collins, CO, USA \\
ali.pezeshki@colostate.edu}
\and
\IEEEauthorblockN{Sandeep Gogineni}
\IEEEauthorblockA{\textit{Information Systems Laboratories Inc.}\\
Dayton, OH, USA \\
sgogineni@islinc.com} \\
\IEEEauthorblockN{Muralidhar Rangaswamy}
\IEEEauthorblockA{\textit{Air Force Research Laboratory} \\
Wright-Patterson AFB, OH, USA \\
muralidhar.rangaswamy@us.af.mil}
\and
\IEEEauthorblockN{Bosung Kang}
\IEEEauthorblockA{\textit{University of Dayton} \\
Dayton, OH, USA \\
bosung.kang@udri.udayton.edu} \\
\IEEEauthorblockN{Vahid Tarokh}
\IEEEauthorblockA{\textit{Duke University} \\
Durham, NC, USA \\
vahid.tarokh@duke.edu}
}

\maketitle

\begin{abstract}
Recent works exploring data-driven approaches to classical problems in adaptive radar have demonstrated promising results pertaining to the task of radar target localization. Via the use of space-time adaptive processing (STAP) techniques and convolutional neural networks, these data-driven approaches to target localization have helped benchmark the performance of neural networks for matched scenarios. However, the thorough bridging of these topics across mismatched scenarios still remains an open problem. As such, in this work, we augment our data-driven approach to radar target localization by performing a subspace perturbation analysis, which allows us to benchmark the localization accuracy of our proposed deep learning framework across mismatched scenarios. To evaluate this framework, we generate comprehensive datasets by randomly placing targets of variable strengths in mismatched constrained areas via RFView\textsuperscript{\tiny\textregistered}, a high-fidelity, site-specific modeling and simulation tool. For the radar returns from these constrained areas, we generate heatmap tensors in range, azimuth, and elevation using the normalized adaptive matched filter (NAMF) test statistic. We estimate target locations from these heatmap tensors using a convolutional neural network, and demonstrate that the predictive performance of our framework in the presence of mismatches can be predetermined.
\end{abstract}


\section{Introduction}
The limiting performance of space-time adaptive processing (STAP) techniques for radar target localization has been extensively detailed in previous data-driven explorations of classical problems in radar STAP (see \cite{Shyam_STAP, shyam_TAES}). Fundamentally, much of these limitations are attributable to underlying assumptions made within classical radar STAP literature that degrade target localization post-STAP detection (see \cite{murali_stap02, melvin_stap96, guerci_stap00}). First among them is that the interference due to clutter --- modeled via the clutter covariance matrix --- must be estimated using available training data. Traditionally, this estimation is performed using a limited number of radar returns, whereby the accuracy of this estimation governs the performance of radar STAP \cite{guerci2003space}. When the dimensionality of the STAP weight vector grows large, so does the amount of data required for obtaining a good estimate of the clutter covariance matrix (see \cite{murali_stap02, melvin_stap96}). Secondly, when a target lies close to the clutter, there exists a large degree of overlap between the azimuth-Doppler subspace of the target response and that of the clutter response. Projecting the clutter response out (through either orthogonal or oblique projections) inadvertently results in a partial removal of the target response, which amplifies the error in localization post-STAP detection.

Developed as a means to address the limiting performance of traditional radar STAP methods, data-driven approaches to these classical problems (see \cite{Shyam_STAP}, \cite{shyam_TAES}) have shown considerable promise, paralleling the emergence of big data in deep learning applications. The rise of these methods has been made possible by the development of simulators that generate highly accurate radar data. The state-of-the-art software for this application, RFView\textsuperscript{\tiny\textregistered}, possesses high-fidelity data generation capabilities, enabling users to synthesize radar returns via a splatter, clutter, and target signal (SCATS) phenomenology engine. Through RFView\textsuperscript{\tiny\textregistered}, it is possible to generate comprehensive datasets for radar STAP applications. For a complete review of RFView\textsuperscript{\tiny\textregistered} and its capabilities, we direct the reader to \cite{gogineni_RFView}.

As this pertains to our procedure, we generate radar returns using RFView\textsuperscript{\tiny\textregistered} for randomized target locations and strengths within a constrained area in the presence of high-fidelity clutter returns. In this constrained area, we produce heatmap tensors using the Normalized Adaptive Matched Filter (NAMF) test statistic in range, azimuth, and elevation angle. Our objective is to predict targets locations from these 3D heatmap tensors using a robust deep learning framework for target localization. In our prior work (see \cite{Shyam_STAP}), we demonstrated that this framework achieves considerable gains over traditional approaches that invoke linear estimation principles. However, much of this analysis was performed in a `matched' setting, where our convolutional neural network (CNN) architecture was trained and validated on heatmap tensors from the same constrained area (the `original platform location' instance in \cite{shyam_TAES}). To benchmark the localization performance of our CNN framework across `mismatched' scenarios --- where network training and testing is performed with tensors from disparate constrained areas --- we separately used the mean normalized output SCNR metric \cite{reed_rapid}. From a hypotheses testing perspective, this metric relies on information presented by the alternative hypothesis (where the target signal is present in each matched filtered radar array data matrix \cite{liu_SINR}) \& is the normalized output SCNR averaged across the $N$ heatmap tensors comprising the dataset. More generally speaking, the mean normalized output SCNR is further related to the perturbation caused by displacing our original platform location instance. This perturbation is quantifiable in terms of the `subspace perturbation error', which can be derived using a computationally inexpensive comparison of the clutter-plus-noise subspaces pertaining to the null hypothesis \cite{shah_dimension}. We aim to benchmark the predictive performance of our CNN via this more generalized metric to demonstrate the interpretability of our deep learning framework across mismatched scenarios.

The outline of the paper is described below. In Section \ref{Sec2}, we introduce RFView\textsuperscript{\tiny\textregistered} and further describe the robust mismatched case RFView\textsuperscript{\tiny\textregistered} example scenario our CNN framework is tested on. This robust scenario is an extension of the mismatched case scenario introduced in \cite{shyam_TAES}. In Section \ref{Sec3}, we review the NAMF test statistic, which is utilized to generate the heatmap tensors for our training and test datasets. We also outline the average Euclidean distance metric for measuring localization accuracy, and the chordal distance metric used to measure the subspace perturbation error. In Section \ref{Sec4}, we summarize our regression CNN framework for target location estimation. In Section \ref{Sec5}, we present numerical results for the robust mismatched case, and demonstrate the feasibility of using the chordal distance as a metric to benchmark network robustness to perturbations. Subsequently, we delineate how few-shot learning can be used to improve network generalization. Finally, in Section \ref{Sec6}, we summarize this work and provide concluding remarks.

\section{RFView\textsuperscript{\tiny\textregistered} Preliminaries} \label{Sec2}
We begin by introducing the RFView\textsuperscript{\tiny\textregistered} modeling and simulation environment, which has been used to construct the robust mismatched case example scenario in this work.

\subsection{RFView\textsuperscript{\tiny\textregistered} Modeling and Simulation Tool} \label{Sec2.1}
RFView\textsuperscript{\tiny\textregistered} \cite{gogineni_RFView} is a high-fidelity site-specific RF modeling and simulation environment developed by ISL Inc. that operates on a world-wide database of terrain and land cover data. With this simulation platform, one can generate representative radar data for use in various signal processing applications. Examples of tools incorporated into RFView\textsuperscript{\tiny\textregistered} include target returns, ground scattered clutter returns, direct path signal, and coherent and incoherent interference effects. To specify simulation scenarios and parameters in RFView\textsuperscript{\tiny\textregistered}, one can use either a MATLAB package or access a cloud-based web interface.

\begin{table}[h!]
\caption{Shared Site and Radar Parameters}
\label{radar parameters}
\centering
\begin{tabularx}{\columnwidth}{s|a}
\hline Parameters & Values \\ \hline
Carrier frequency & $10,000 \ \mathrm{MHz}$ \\
Bandwidth & $5 \ \mathrm{MHz}$ \\
PRF \& Duty Factor & $1100 \ \mathrm{Hz}$ \& 10 \\
Receiving antenna & $48 \times 5$ (horizontal $\times$ vertical elements) \\
Transmitting antenna & $48 \times 5$ (horizontal $\times$ vertical elements) \\
Antenna element spacing & $0.015 \ \mathrm{m}$ \\
Platform height & $1000 \ \mathrm{m}$ \\
Area latitude (min, max) & $(32.4611^{\circ},32.6399^{\circ})$\\
Area longitude (min, max) & $(-117.1554^{\circ},-116.9433^{\circ})$\\
\end{tabularx}
\end{table}

By utilizing these functionalities, one can define synthetic example scenarios within RFView\textsuperscript{\tiny\textregistered} that accurately model real-world environments. Our considered example scenario consists of a stationary airborne radar platform system flying over the coast of Southern California. The simulation region covers a $20 \ \text{km} \times 20 \ \text{km}$ area, as shown in Figure \ref{mismatched map}. RFView\textsuperscript{\tiny\textregistered} aggregates the information on land types, the geographical characteristics across the simulation region, and the site and radar parameters to simulate the radar return signal. These radar parameters are given in Table \ref{radar parameters}---we consider a single-channel transmitter and an $L$-channel receiver, where the size-$(48 \times 5)$ receiver array is condensed to size-$(L \times 1)$. Subsequently, the radar return is beamformed for each size-$(48/L \times 5)$ receiver sub-array. The radar operates in `spotlight' mode and points toward the center of the simulation region, where each radar return is produced using a single transmitted pulse.

\subsection{Robust Mismatched Case RFView\textsuperscript{\tiny\textregistered} Example Scenario} \label{Sec2.2}
\begin{figure*}[hbt!]
    \centering
    \includegraphics[width=0.68\linewidth]{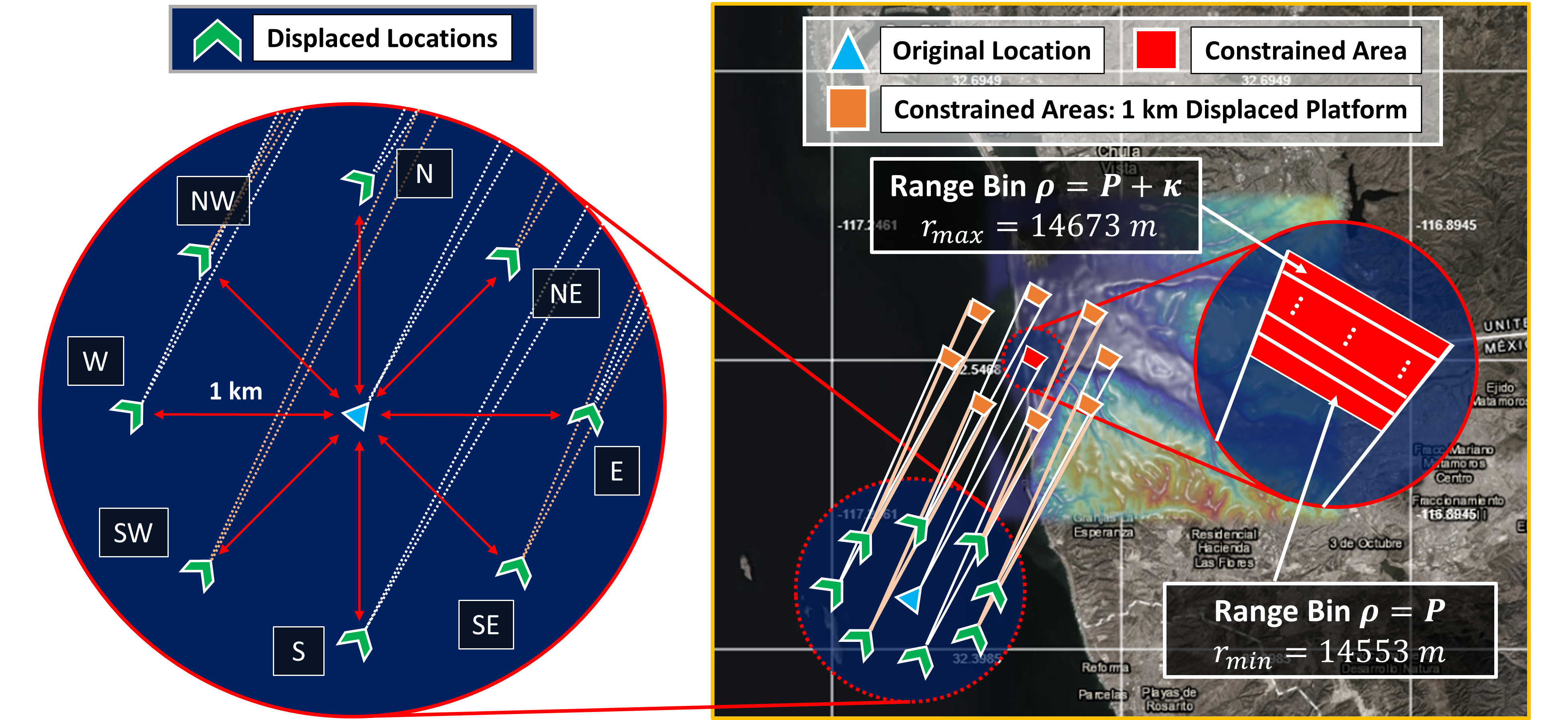}
    \caption{The map of the robust mismatched case RFView\textsuperscript{\tiny\textregistered} example scenario. Each of the displaced platform locations (N, NW, W, SW, S, SE, E, and NE) are shown in green, and their relative constrained areas for target placement are depicted in orange. The blue triangle represents the original platform location and the red region is its respective constrained area for target placement. The elevation heatmap overlaying the right image represents the simulation region.}
    \label{mismatched map}
\end{figure*}

For our robust mismatched case RFView\textsuperscript{\tiny\textregistered} example scenario, we first consider an airborne radar platform within the scene parameterized by Table \ref{radar parameters}: \textbf{(O)} (the original platform location). Next, we consider a $1$ km displacement, \textbf{(D)}, of this airborne radar platform in each cardinal and intermediate direction: \textbf{(N)}, \textbf{(NW)}, \textbf{(W)}, \textbf{(SW)}, \textbf{(S)}, \textbf{(SE)}, \textbf{(E)}, and \textbf{(NE)}. For each of these platform location instances, we randomly place a single target within a corresponding constrained area that contains $\kappa$ range bins and varies in range, $r$, azimuth, $\theta$, and elevation, $\phi$, where:
\begin{align}
\begin{split}
    &r \in [r_{\mathrm{min}}-\Delta r/2,r_{\mathrm{max}} + \Delta r/2] \\
    \theta &\in [\theta_{\mathrm{min}},\theta_{\mathrm{max}}], \quad \phi \in [\phi_{\mathrm{min}},\phi_{\mathrm{max}}]
\end{split}
\end{align}
We note that in RFView\textsuperscript{\tiny\textregistered}, range bin $\rho \in \mathbb{N}$ is defined for:
\begin{align}
    r \in \Big[r_\rho-\frac{\Delta r}{2},r_\rho+\frac{\Delta r}{2}\Big] = [G + \rho \Delta r,G + (\rho+1) \Delta r]
\end{align}
where $r_\rho$ is the midpoint (in meters) of range bin $\rho$, and $G$ is the distance between the platform location and the simulation region. As such, $r_{\mathrm{min}}$ is the midpoint of range bin $\rho = \mathit{P}$ and $r_{\mathrm{max}}$ is the midpoint of range bin $\rho = \mathit{P} + \kappa$ (see Figure \ref{mismatched map}). The default grid resolution is defined as $(\Delta r, \Delta\theta, \Delta\phi)$, where $\Delta r$ denotes the chip size. The target RCS, $\sigma$, is arbitrarily selected from a uniform distribution with specified mean, $\mu$, and range, $l$, such that $\sigma \sim U(\mu-l/2,\mu+l/2)$. 

As such, for the original platform location instance, \textbf{(O)}, we consider $N$ different point targets (each with unique location \& RCS) and generate $K$ radar returns for each using RFView\textsuperscript{\tiny\textregistered}, yielding ($N \times \kappa$) matched filtered radar array data matrices. Subsequently, for each of the eight displaced platform location instances, we consider $0.1N$ different point targets (each with unique location \& RCS) and generate $K$ radar returns for each, yielding ($0.1N \times \kappa$) matched filtered radar array data matrices for each \textbf{(D)}. The simulation parameters are provided in Table \ref{mismatched parameters}, and the platform locations are displayed in Figure \ref{mismatched map}.

\begin{table}[hbt!]
\caption{Robust Mismatched Case Site and Radar Parameters}
\label{mismatched parameters}
\centering
\fontsize{8}{8.7}\selectfont
\begin{tabularx}{\columnwidth}{b|t}
\hline \textbf{Original Location (O)} -- Parameters & Values \\ \hline
Platform latitude, longitude & $32.4275^{\circ},-117.1993^{\circ}$\\
Constrained area range $(r_{\mathrm{min}},r_{\mathrm{max}})$ & $(14553 \ \mathrm{m},14673 \ \mathrm{m})$ \\
Constrained area azimuth $(\theta_{\mathrm{min}},\theta_{\mathrm{max}})$ & $(20^{\circ},30^{\circ})$ \\
Constrained area elevation $(\phi_{\mathrm{min}},\phi_{\mathrm{max}})$ & $(-4.1^{\circ},-3.9^{\circ})$ \\
\hline \textbf{1 km North (N)} -- Parameters & Values \\ \hline
Platform latitude, longitude & $32.4095^{\circ},-117.1993^{\circ}$\\
Constrained area range $(r_{\mathrm{min}},r_{\mathrm{max}})$ & $(13800 \ \mathrm{m},13920 \ \mathrm{m})$ \\
Constrained area azimuth $(\theta_{\mathrm{min}},\theta_{\mathrm{max}})$ & $(20^{\circ},30^{\circ})$ \\
Constrained area elevation $(\phi_{\mathrm{min}},\phi_{\mathrm{max}})$ & $(-4.3^{\circ},-4.1^{\circ})$ \\
\hline \textbf{1 km Northwest (NW)} -- Parameters & Values \\ \hline
Platform latitude, longitude & $32.4070^{\circ},-117.1920^{\circ}$\\
Constrained area range $(r_{\mathrm{min}},r_{\mathrm{max}})$ & $(14467 \ \mathrm{m},14587 \ \mathrm{m})$ \\
Constrained area azimuth $(\theta_{\mathrm{min}},\theta_{\mathrm{max}})$ & $(20^{\circ},30^{\circ})$ \\
Constrained area elevation $(\phi_{\mathrm{min}},\phi_{\mathrm{max}})$ & $(-4.1^{\circ},-3.9^{\circ})$ \\
\hline \textbf{1 km West (W)} -- Parameters & Values \\ \hline
Platform latitude, longitude & $32.4005^{\circ},-117.2099^{\circ}$\\
Constrained area range $(r_{\mathrm{min}},r_{\mathrm{max}})$ & $(15207 \ \mathrm{m},15327 \ \mathrm{m})$ \\
Constrained area azimuth $(\theta_{\mathrm{min}},\theta_{\mathrm{max}})$ & $(20^{\circ},30^{\circ})$ \\
Constrained area elevation $(\phi_{\mathrm{min}},\phi_{\mathrm{max}})$ & $(-3.95^{\circ},-3.75^{\circ})$ \\
\hline \textbf{1 km Southwest (SW)} -- Parameters & Values \\ \hline
Platform latitude, longitude & $32.3940^{\circ},-117.2066^{\circ}$\\
Constrained area range $(r_{\mathrm{min}},r_{\mathrm{max}})$ & $(15544 \ \mathrm{m},15664 \ \mathrm)$ \\
Constrained area azimuth $(\theta_{\mathrm{min}},\theta_{\mathrm{max}})$ & $(20^{\circ},30^{\circ})$ \\
Constrained area elevation $(\phi_{\mathrm{min}},\phi_{\mathrm{max}})$ & $(-3.85^{\circ},-3.65^{\circ})$ \\
\hline \textbf{1 km South (S)} -- Parameters & Values \\ \hline
Platform latitude, longitude & $32.3915^{\circ},-117.1993^{\circ}$\\
Constrained area range $(r_{\mathrm{min}},r_{\mathrm{max}})$ & $(15321 \ \mathrm{m},15441 \ \mathrm{m})$ \\
Constrained area azimuth $(\theta_{\mathrm{min}},\theta_{\mathrm{max}})$ & $(20^{\circ},30^{\circ})$ \\
Constrained area elevation $(\phi_{\mathrm{min}},\phi_{\mathrm{max}})$ & $(-3.9^{\circ},-3.7^{\circ})$ \\
\hline \textbf{1 km Southeast (SE)} -- Parameters & Values \\ \hline
Platform latitude, longitude & $32.3940^{\circ},-117.1920^{\circ}$\\
Constrained area range $(r_{\mathrm{min}},r_{\mathrm{max}})$ & $(14680 \ \mathrm{m},14800 \ \mathrm{m})$ \\
Constrained area azimuth $(\theta_{\mathrm{min}},\theta_{\mathrm{max}})$ & $(20^{\circ},30^{\circ})$ \\
Constrained area elevation $(\phi_{\mathrm{min}},\phi_{\mathrm{max}})$ & $(-4.05^{\circ},-3.85^{\circ})$ \\
\hline \textbf{1 km East (E)} -- Parameters & Values \\ \hline
Platform latitude, longitude & $32.4005^{\circ},-117.1887^{\circ}$\\
Constrained area range $(r_{\mathrm{min}},r_{\mathrm{max}})$ & $(13921 \ \mathrm{m},14041 \ \mathrm{m})$ \\
Constrained area azimuth $(\theta_{\mathrm{min}},\theta_{\mathrm{max}})$ & $(20^{\circ},30^{\circ})$ \\
Constrained area elevation $(\phi_{\mathrm{min}},\phi_{\mathrm{max}})$ & $(-4.25^{\circ},-4.05^{\circ})$ \\
\hline \textbf{1 km Northeast (NE)} -- Parameters & Values \\ \hline
Platform latitude, longitude & $32.4070^{\circ},-117.1920^{\circ}$\\
Constrained area range $(r_{\mathrm{min}},r_{\mathrm{max}})$ & $(13558 \ \mathrm{m},13678 \ \mathrm{m})$ \\
Constrained area azimuth $(\theta_{\mathrm{min}},\theta_{\mathrm{max}})$ & $(20^{\circ},30^{\circ})$ \\
Constrained area elevation $(\phi_{\mathrm{min}},\phi_{\mathrm{max}})$ & $(-4.35^{\circ},-4.15^{\circ})$ \\
\end{tabularx}
\end{table}

\section{Heatmap Tensor Generation and Benchmarking Metrics} \label{Sec3}
In our analysis, we make use of the NAMF test statistic \cite{michels_performance} to generate heatmap tensors in range, azimuth, and elevation of the constrained area(s) pertaining to our robust mismatched case RFView\textsuperscript{\tiny\textregistered} example scenario (see Section \ref{Sec2.2}). The signal model for our localization framework is derived from the radar STAP detection problem (hypothesis testing on radar returns). However, as we know that each radar return contains a single point target (see Section \ref{Sec3}), we solely consider the post-STAP detection problem of target localization.

\subsection{Heatmap Tensor Generation} \label{Sec3.1}
We first consider an $L$-element receiver array for the radar receiver. Let $\mathbf{Y_\rho}\in \bbC^{L \times K}$ be a matrix comprising $K$ independent radar returns (radar array data) and let $\mathbf{Z_\rho}\in \bbC^{L \times K}$ be a matrix comprising $K$ independent realizations of the clutter-plus-noise data, both of which have been matched filtered to range bin $\rho \in \{P,P+1,...,P+\kappa\}$, where $r_\rho = G + \rho\Delta r + \Delta r/2$ is the midpoint of range bin $\rho$, and $G$ is the distance between the platform location and the simulation region. We now define $\mathit{H}_0$ and $\mathit{H}_1$ as the null and alternative hypotheses, respectively. We can derive the following signal model, where $\mathbf{X_\rho}\in \bbC^{L \times K}$ consists of $K$ independent realizations of the matched filtered target data, $\mathbf{C_\rho}, \mathbf{\bar{C}_\rho} \in \bbC^{L \times K}$ are unique sets of $K$ independent realizations of the matched filtered clutter data, and $\mathbf{N_\rho}, \mathbf{\bar{N}_\rho}\in \bbC^{L \times K}$ are unique sets of $K$ independent realizations of the matched filtered noise data\footnotemark:
\begin{align}
    &\mathit{H}_0: \mathbf{Z_\rho} = \mathbf{\overbar{C}_\rho} + \mathbf{\overbar{N}_\rho} \\
    &\mathit{H}_1: \mathbf{Y_\rho} = \mathbf{X_\rho} + \mathbf{C_\rho} + \mathbf{N_\rho} \label{eq_alternative}
\end{align}

\footnotetext{\label{note1}We note that all applicable data matrices (e.g., $\mathbf{Z_\rho},\mathbf{Y_\rho}$) are preliminarily mean centered ($\mathbb{E}[\mathbf{Z_\rho}] = \mathbb{E}[\mathbf{Y_\rho}] = \vec{0}$) before subsequent data processing.}

\noindent Let $\mathbf{a_\rho}(\theta, \phi) \in \mathbb{C}^L$ denote the array steering vector associated with coordinates $(r_\rho, \theta, \phi)$ in range, azimuth and elevation (the array steering vector is provided by RFView\textsuperscript{\tiny\textregistered}). Subsequently, we whiten the array steering vector alongside the clutter-plus-noise component of $\mathbf{Y_\rho}$, the matched filtered radar array data:
\begin{align}
    &\mathbf{\hat{Y}_\rho} = \Sigma^{-1/2}\mathbf{Y_\rho}, \quad \ \mathbf{\hat{a}_\rho}(\theta, \phi) = \Sigma^{-1/2}\mathbf{a_\rho}(\theta, \phi) \\
    &\textbf{where:} \ \ \Sigma = \mathbb{E}[\mathbf{Z_\rho}\mathbf{Z_\rho}^H] \label{eq_covariance}
\end{align}

\noindent Using this $\mathbf{\hat{Y}_\rho}$ \& $\mathbf{\hat{a}_\rho}(\theta, \phi)$, the NAMF test statistic, $\Gamma_\rho(\theta,\phi) \in \mathbb{R}^+$ (derived from \cite{michels_performance}), for coordinates $(r_\rho, \theta, \phi)$ is given by:
\begin{align}
    \Gamma_\rho(\theta,\phi)= \frac{\| \mathbf{\hat{a}_\rho}(\theta, \phi)^H \mathbf{\hat{Y}_\rho}\|_2^2}{[\mathbf{\hat{a}_\rho}(\theta, \phi)^H \mathbf{\hat{a}_\rho}(\theta, \phi)] \ \|\text{diag}(\mathbf{\hat{Y}_\rho}^H \mathbf{\hat{Y}_\rho})\|_2}
\end{align}

\noindent Thereupon, sweeping the steering vector across $\theta$ and $\phi$ at the default angular resolution, $(\Delta \theta,\Delta \phi)$, and recording $\Gamma_\rho(\theta,\phi)$ at each location, we produce a heatmap image in azimuth and elevation. Stacking these images over the $\kappa$ consecutive range bins (indexed by $\rho$) produces a 3-dimensional heatmap tensor for STAP; these heatmap tensors comprise the examples of our dataset. Per Section \ref{Sec2.2}, we have $N \times \kappa$ (original) \& $0.1N \times \kappa$ (displaced) matched filtered radar array data matrices, which produce $N$ \& $0.1N$ heatmap tensors, respectively.

Through the matched filtering and whitening transformation procedures described above, the SCNR is much increased. This improved ratio is defined as the output SCNR, which is unique for each of the $\kappa$ range bins that constitute a given heatmap tensor. From Eq. (\ref{eq_alternative}), the output SCNR is given by\textsuperscript{\ref{note1}}:
\begin{align}
    &(\text{SCNR}_{\text{Output}})_\rho = 10\log_{10} \left[ \frac{\Tr(\mathbf{X_\rho}^H\Sigma^{-1}\mathbf{X_\rho})}{\Tr(\mathbf{W_\rho}^H\Sigma^{-1}\mathbf{W_\rho})} \right] \\
    &\textbf{where:}  \ \ \mathbf{W_\rho} = \mathbf{C_\rho} + \mathbf{N_\rho}
\end{align}

\noindent We recall (per Section \ref{Sec2.2}) that for each of our $N$ (original) heatmap tensors, a single target is positioned in one of the $\kappa$ range bins ($\rho = \hat{\rho}$). Thus, we only consider $(\text{SCNR}_{\text{Output}})_{\rho}$ for $\rho = \hat{\rho}$. The mean of these $N$ (original) $(\text{SCNR}_{\text{Output}})_{\hat{\rho}}$ values yields the \textbf{Mean Output SCNR} of our dataset (see Eq. (\ref{eq_SCNR})). Our empirical results are conveyed through this measure.
\begin{equation} \label{eq_SCNR}
    (\overline{\text{SCNR}}_{\text{Output}})_{\hat{\rho}} = \frac{1}{N} \sum_{i=1}^N \left[(\text{SCNR}_\text{Output})_{\hat{\rho}}\right]_i
\end{equation}

We record the ground truth target location for each heatmap tensor example using the standard Cartesian coordinate system, with the platform at the origin, the Northward-pointing line as the x-axis, and the upward-pointing line as the z-axis. Our final dataset comprises the heatmap tensors as the features and the coordinate encoded true target locations (see Section VI.B in \cite{shyam_TAES}) as the labels, which will be used by our CNN.

\subsection{Average Euclidean Distance Metric} \label{Sec3.2}
Following the heatmap tensor generation process outlined in Section \ref{Sec3.1}, we must now define an error metric to interpret and compare the localization accuracies of our NAMF test statistic and augmented CNN frameworks. While our heatmap tensors follow the spherical coordinate system, for our definition of the localization accuracy, we transform the ground truth target location into Cartesian coordinates. Moreover, to report the localization error in meters, we use the \textbf{Average Euclidean Distance} between the predicted target locations and the ground truth target locations to plot the localization error. As such, we let $(r_i,\theta_i,\phi_i) \rightarrow (x_i,y_i,z_i)$ represent the ground truth target location for example $i$ from our dataset and let $(\thickhat{x}_i, \thickhat{y}_i, \thickhat{z}_i)$ be the estimated target location for this example (predicted by the CNN model). The localization error, $Err_{\text{CNN}}$, over the $N_{test}$ test examples, is defined as:
\begin{align}
Err_{\text{CNN}} = \frac{\sum_{i = 1}^{N_{test}} \| (x_i, y_i, z_i) - (\thickhat{x}_i, \thickhat{y}_i, \thickhat{z}_i)\|_2}{N_{test}}
\end{align}

We compare this target localization error from our regression network with the error from a (more) traditional approach of using the midpoint of the grid cell with the peak NAMF test statistic. Let $(\thickbar{r}_i, \thickbar{\theta}_i, \thickbar{\phi}_i) \rightarrow (\thickbar{x}_i, \thickbar{y}_i, \thickbar{z}_i)$ be the midpoint of the peak grid cell for example $i$ from the particular dataset. Over the $N_{test}$ test examples, we can compute the error, $Err_{\text{NAMF}}$, in using this (more) classical approach as:
\begin{align}
Err_{\text{NAMF}} = \frac{\sum_{i = 1}^{N_{test}} \| (x_i, y_i, z_i) - (\thickbar{x}_i, \thickbar{y}_i, \thickbar{z}_i)\|_2}{N_{test}}
\end{align}

\subsection{Chordal Distance Metric} \label{Sec3.3}
Among the key challenges of deep neural networks is that their generalization capabilities cannot be trivially estimated prior to evaluation. In our approach, the term `generalization' pertains to the evaluation of our CNN architecture (see Section \ref{Sec4}) on the displaced platform location instances, having been trained on the original platform location instance. However, extending from the null hypothesis presented in Section \ref{Sec3.1}, we see that a measure of similarity between the clutter (clutter-plus-noise) subspaces (see \cite{pezeshki_empirical}) of the original and displaced platform location instances can be used to preliminarily gauge network generalization. We outline this similarity measure via the chordal distance metric for the subspace perturbation error, first detailed by Shah and Tufts in the context of estimating the dimensionality of noisy signal subspaces \cite{shah_dimension}.

Following the outline in Section \ref{Sec3}, we consider the clutter-plus-noise data matrices $\mathbf{Z_\rho}^\textbf{{(O)}}, \mathbf{Z_\rho}^\textbf{{(D)}} \in \bbC^{L \times K}$ for the original and displaced platform location instances, respectively, where:
\begin{gather}
    \textbf{(D)} \in \{\textbf{N,\ NW,\ W,\ SW,\ S,\ SE,\ E,\ NE}\} \\
    \mathbf{Z_\rho}^\textbf{(O)} = \mathbf{\overbar{C}_\rho}^\textbf{(O)} + \mathbf{\overbar{N}_\rho}^\textbf{(O)} \\
    \mathbf{Z_\rho}^\textbf{(D)} = \mathbf{\overbar{C}_\rho}^\textbf{(D)} + \mathbf{\overbar{N}_\rho}^\textbf{(D)}
\end{gather}
To identify the clutter subspaces for \textbf{(O)} \& \textbf{(D)}, we must determine the rank of the clutter-only data matrices, $\mathbf{\overbar{C}_\rho}^\textbf{{(O)}}, \mathbf{\overbar{C}_\rho}^\textbf{{(D)}} \in \bbC^{L \times K}$, which we approximate by performing a singular value decomposition (SVD) of the clutter-plus-noise covariance matrices, $\Sigma^\textbf{{(O)}} = \mathbb{E}[\mathbf{Z_\rho}^\textbf{(O)}\mathbf{Z_\rho}^{\textbf{(O)}H}]$ and $\Sigma^\textbf{{(D)}} = \mathbb{E}[\mathbf{Z_\rho}^\textbf{(D)}\mathbf{Z_\rho}^{\textbf{(D)}H}]$.
\begin{align}
    &\Sigma^\textbf{(O)} = \mathbf{U_{\overbar{C}}}^\textbf{(O)} \, \mathbf{S_{\overbar{C}}}^\textbf{(O)} \, \mathbf{U_{\overbar{C}}}^{\textbf{(O)}H} + \mathbf{U_{\overbar{N}}}^\textbf{(O)} \, \mathbf{S_{\overbar{N}}}^\textbf{(O)} \, \mathbf{U_{\overbar{N}}}^{\textbf{(O)}H} \\
    &\Sigma^\textbf{(D)} = \mathbf{U_{\overbar{C}}}^\textbf{(D)} \, \mathbf{S_{\overbar{C}}}^\textbf{(D)} \, \mathbf{U_{\overbar{C}}}^{\textbf{(D)}H} + \mathbf{U_{\overbar{N}}}^\textbf{(D)} \, \mathbf{S_{\overbar{N}}}^\textbf{(D)} \, \mathbf{U_{\overbar{N}}}^{\textbf{(D)}H}
\end{align}
We note that $\mathbf{U_{\overbar{C}}}^\textbf{(O)}, \mathbf{U_{\overbar{C}}}^\textbf{(D)} \in \bbC^{L \times r}$ contain the first $\boldsymbol{r}$ singular vectors of $\Sigma^\textbf{(O)}$ and $\Sigma^\textbf{(D)}$, which span the approximated column spaces of $\mathbf{\overbar{C}_\rho}^\textbf{{(O)}}$ and $\mathbf{\overbar{C}_\rho}^\textbf{{(D)}}$, respectively. Subsequently, we can define the estimated clutter-only symmetric complex matrices, $\mathbf{J_{\overbar{C}}}^\textbf{(O)}, \mathbf{J_{\overbar{C}}}^\textbf{(D)} \in \bbC^{L \times L}$, with which we approximate the chordal distance, $\mathbf{\hat{d}}_{\text{chordal}}$, between $\mathbf{\overbar{C}_\rho}^\textbf{{(O)}}$ and $\mathbf{\overbar{C}_\rho}^\textbf{{(D)}}$ \cite{Li1991PerformanceAO}.
\begin{align}
    &\mathbf{J_{\overbar{C}}}^\textbf{(O)} = \mathbf{U_{\overbar{C}}}^\textbf{(O)} \, \mathbf{U_{\overbar{C}}}^{\textbf{(O)}H}, \quad
    \mathbf{J_{\overbar{C}}}^\textbf{(D)} = \mathbf{U_{\overbar{C}}}^\textbf{(D)} \, \mathbf{U_{\overbar{C}}}^{\textbf{(D)}H} \\
    &\mathbf{\hat{d}}_{\text{chordal}} = \boldsymbol{r} - \Tr(\mathbf{J_{\overbar{C}}}^\textbf{(D)} \, \mathbf{J_{\overbar{C}}}^\textbf{(O)} \, \mathbf{J_{\overbar{C}}}^\textbf{(D)})
\end{align}
Above, $\mathbf{\hat{d}}_{\text{chordal}}$ is a measure of the subspace perturbation error, and thus measures the similarity between \textbf{(O)} and \textbf{(D)}.

\section{Deep Learning Framework} \label{Sec4}
\begin{figure}[h!]
    \centering
    \includegraphics[width=1\linewidth]{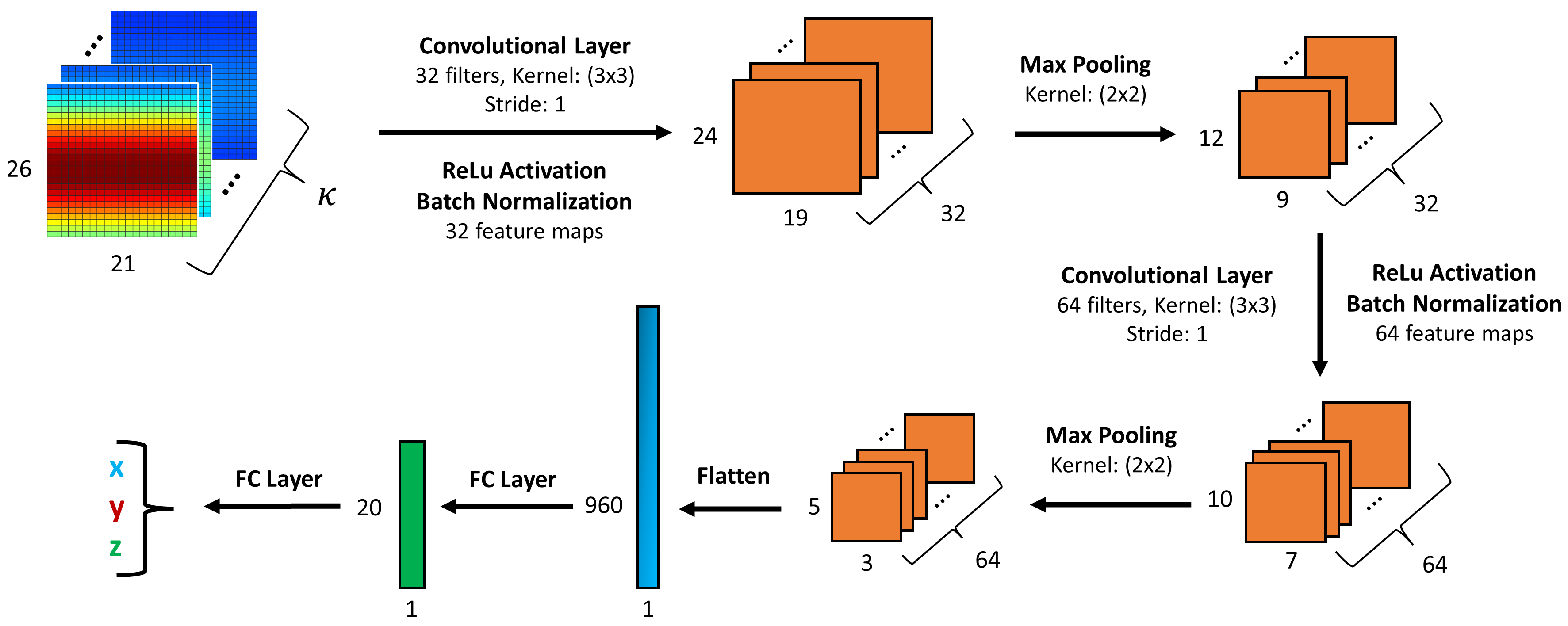}
    \caption{CNN architecture for $(\Delta r,\Delta \theta,\Delta \phi) = (30 \ \text{m},0.4^{\circ},0.01^{\circ})$.}
    \label{CNN_default_parameters}
\end{figure}

We reconsider the regression CNN framework proposed in \cite{Shyam_STAP} (see Figure \ref{CNN_default_parameters}) to estimate the position of a single target in the presence of clutter and noise using heatmap tensors from the NAMF test statistic. For each step in these evaluations, we produce a training dataset consisting of $N$ heatmap tensors (original location) and eight test datasets consisting of $0.1N$ heatmap tensors (displaced locations). Each of our $N$ training examples and $0.1N$ test examples are of size $\kappa \times 26 \times 21$, and can be visualized as an array of $\kappa$ heatmaps (one for each range bin), each of size $26 \times 21$ ($\theta$-dimension $\times$ $ \phi$-dimension). For a complete review of this CNN, we refer the reader to \cite{shyam_TAES}.

\begin{figure*}[h!]
    \centering
    \includegraphics[width=0.92\linewidth]{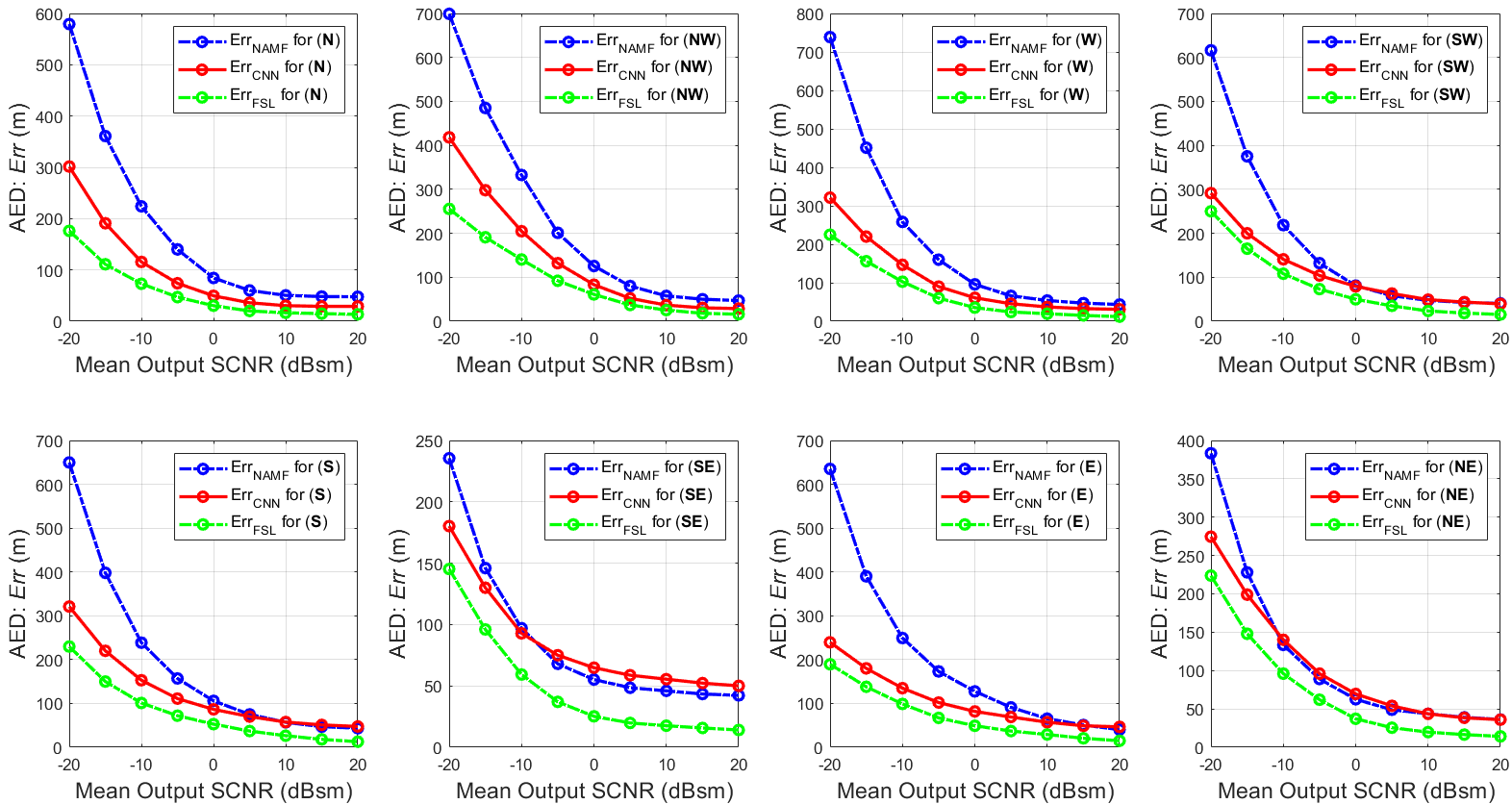}
    \caption{Comparing target localization error (average Euclidean distance [\textbf{AED}]) of the CNN model from Section \ref{Sec4} (in red) with the traditional cell-midpoint method (in blue) using the NAMF test statistic for each of the displaced platform location instances: \textbf{(N)}, \textbf{(NW)}, \textbf{(W)}, \textbf{(SW)}, \textbf{(S)}, \textbf{(SE)}, \textbf{(E)}, and \textbf{(NE)}. The CNN model is trained on the original platform location instance, \textbf{(O)}, and is augmented with few-shot learning for the FSL case.}
    \label{Gain_vs_RCS}
\end{figure*}

\section{Robust Mismatched Case Empirical Results} \label{Sec5}
As outlined in Section \ref{Sec4}, we evaluate our regression CNN framework in the mismatched case to estimate the position of a single point target in the presence of clutter and noise using heatmap tensors from the NAMF test statistic. We gauge how well our framework can generalize to these new scenarios, and augment our analysis via few-shot learning.

\subsection{Evaluating CNN Framework for Robust Mismatched Case} \label{Sec5.1}
For our analysis, we consider the scenario outlined in Section \ref{Sec2.2} over a range of SCNR values, where $(\overline{\text{SCNR}}_{\text{Output}})_{\hat{\rho}} \in [-20 \ \text{dB},20 \ \text{dB}]$, with $\Delta(\overline{\text{SCNR}}_{\text{Output}})_{\hat{\rho}} = 5 \ \text{dB}$. Furthermore, we let L = 16, K = 100, and now consider the site and radar parameters provided in Table \ref{mismatched parameters}, with default grid resolution $(\Delta r,\Delta\theta,\Delta\phi) = (30 \ \text{m}, 0.4^{\circ}, 0.01^{\circ})$ such that $\kappa = 5$. Next, for each $(\overline{\text{SCNR}}_{\text{Output}})_{\hat{\rho}}$ value, we produce one training dataset consisting of $N = 9 \times 10^4$ heatmap tensors and test datasets consisting of $0.1N = 9 \times 10^3$ heatmap tensors, for the original and the displaced platform location instances, respectively. We now have one training dataset (derived for \textbf{(O)}), and eight test datasets (one for each displaced platform location instance).

Subsequently, we consider the CNN from Section \ref{Sec4}, and train this network on the original platform location instance using the Adam optimizer with $\alpha = 1 \times 10^{-3}$ \cite{kingma_14}. We apply this network to the displaced platform location instances, and compare the resulting target localization accuracies with the (more) classical approach, using the metric from Section \ref{Sec3.2}. The results of this analysis are depicted in Figure \ref{Gain_vs_RCS}.

Recalling the chordal distance metric for measuring scenario similarity (see Section \ref{Sec3.3}), Table \ref{chordal distance} summarizes the chordal distances, $\mathbf{\hat{d}}_{\text{chordal}}$, between the original platform location instance and each displaced platform location instance. The gain afforded by our CNN (over the (more) classical approach) is also provided for each displaced platform location instance at $(\overline{\text{SCNR}}_{\text{Output}})_{\hat{\rho}} = 20 \ \text{dB}$. The chordal distance informs us that our CNN should achieve the highest gain for the \textbf{(N)}, \textbf{(NW)}, and \textbf{(W)} cases, followed by the \textbf{(SW)}, and \textbf{(NE)} cases. The \textbf{(S)}, \textbf{(E)}, and \textbf{(SE)} cases are expected to yield the lowest gain. We observe that this ordering yields a pairwise agreement with the gain afforded by the CNN (above the breakdown threshold of the NAMF test statistic \cite{shyam_TAES}) at $(\overline{\text{SCNR}}_{\text{Output}})_{\hat{\rho}} = 20 \ \text{dB}$.
\begin{table}[h!]
\caption{Chordal Distance and CNN Gain}
\label{chordal distance}
\centering
\bgroup
\def\arraystretch{1.4}%
\begin{tabular}{c|c|c|c|c|c|c|c|c} 
\hline
\textbf{D} & \textbf{N} & \textbf{NW} & \textbf{W} & \textbf{SW} & \textbf{S} & \textbf{SE} & \textbf{E} & \textbf{NE} \\
\hline
$\mathbf{\hat{d}}_{\text{chordal}}$ & 0.31 & 0.31 & 0.34 & 0.45 & 0.51 & 0.55 & 0.54 & 0.47 \\
\hline
Gain & 1.65 & 1.63 & 1.41 & 1.02 & 0.91 & 0.84 & 0.86 & 1.01
\end{tabular}
\egroup
\end{table}

\subsection{Few-Shot Learning for Robust Mismatched Case} \label{Sec5.2}
Revisiting the mismatched case from Section \ref{Sec5.1}, we note that while our CNN presents improved localization accuracies over the (more) classical method when $\mathbf{\hat{d}}_{\text{chordal}}$ is minimized, the gain is substantially reduced. To ameliorate this reduction, we use few-shot learning (FSL): a method of fine-tuning neural networks trained on large baseline datasets to perform similar tasks using minimal new examples. The complete motivation behind and review of few-shot learning are provided in \cite{shyam_TAES}.

To augment our trained CNN from Section \ref{Sec5.1} with few-shot learning, we freeze the convolutional and batch normalization layers of our trained CNN (the weights and biases will remain constant), halving the number of trainable parameters. Regarding the remaining layers, the weights and biases will be updated via fine-tuning with FSL. Subsequently, we generate a unique dataset for each displaced platform location instance consisting of 64 new heatmap tensors, using the same site and radar parameters specified in Table \ref{mismatched parameters}. Finally, we use few-shot learning to fine-tune this network for each of the displaced platform location instances by further training our CNN using the 64 new examples. We consider the Adam optimizer with reduced learning rate $\alpha = 5 \times 10^{-4}$. The results of this analysis are shown in Figure \ref{Gain_vs_RCS} alongside the results from Section \ref{Sec5.1}.

Using few-shot learning, we observe a consistent improvement in the gain afforded by our trained CNN over the (more) traditional method for all eight displaced platform location instances, with a 3-fold improvement being achieved across all cases for $(\overline{\text{SCNR}}_{\text{Output}})_{\hat{\rho}} = 20 \ \text{dB}$. This analysis demonstrates that our regression CNN framework can be augmented with few-shot learning to improve network generalization and ameliorate the reduced target localization accuracies we observed in Section \ref{Sec5.1} across displaced platform location instances.

\section{Conclusion} \label{Sec6}
While previous data-driven approaches to radar target localization have helped benchmark neural network performance across matched scenarios, the comprehensive bridging of these topics across mismatched scenarios has been an open problem. Consequently, in this work, we benchmarked the performance of our deep learning framework for target localization across mismatched scenarios through a subspace perturbation analysis. Using our CNN architecture and enabled by this analysis, we showed that the predictive performance of our framework in the presence of perturbations could be predetermined. We also showed that the gains afforded by our CNN model across mismatched scenarios could be improved using FSL.

\section*{Acknowledgment}
This work is supported in part by the Air Force Office of Scientific Research under award FA9550-21-1-0235. Dr. Muralidhar Rangaswamy and Dr. Bosung Kang are supported by the AFOSR under project 20RYCORO51. Dr. Sandeep Gogineni is supported by the AFOSR under project 20RYCOR052.

\bibliographystyle{IEEEtran}
\bibliography{conference_101719}

\end{document}